\documentclass[11pt,a4paper]{article}
\usepackage{CJKutf8}
\usepackage[]{acl2020}
\usepackage{times}
\usepackage{amsmath}
\usepackage{latexsym}
\usepackage{mathtools}
\usepackage{mathabx}

\usepackage{subcaption}
\usepackage{tikz}
\usepackage{xspace}
\usetikzlibrary{shapes.geometric, arrows}
\newcommand{\ie}{\textit{i}.\textit{e}.}
\newcommand{\eg}{\textit{e}.\textit{g}.}
\newcommand{\mymethod}{\textsc{CIAT}\xspace}

\usepackage{mwe}
\usepackage{paralist}

\usepackage{microtype}

\title{Counter-Interference Adapter for Multilingual Machine Translation}

\author{Yaoming Zhu\textsuperscript{1} \qquad Jiangtao Feng\textsuperscript{1}\qquad Chengqi Zhao\textsuperscript{1}\qquad Mingxuan Wang\textsuperscript{1}\qquad Lei Li\textsuperscript{2}\thanks{~~Work is done while at ByteDance.} \\ \textsuperscript{1} ByteDance AI Lab\\ \textsuperscript{2} University of California, Santa Barbara \\
\small \texttt{\{zhuyaoming,fengjiangtao,zhaochengqi.d, wangmingxuan.89\}@bytedance.com} \\
\small \texttt{lilei@cs.ucsb.edu} 
}

\date{}

\begin{document}
\begin{CJK}{UTF8}{gbsn}

\maketitle
\begin{abstract}
Developing a unified multilingual model has long been a pursuit for machine translation. 
However, existing approaches suffer from performance degradation --- a single multilingual model is inferior to separately trained bilingual ones on rich-resource languages. 
We conjecture that such a phenomenon is due to interference caused by joint training with multiple languages. 
To accommodate the issue, we propose CIAT, an adapted Transformer model with a small parameter overhead for multilingual machine translation. We evaluate CIAT on multiple benchmark datasets, including IWSLT, OPUS-100, and WMT. 
Experiments show that CIAT consistently outperforms strong multilingual baselines on 64 of total 66 language directions, 42 of which see above 0.5 BLEU improvement. Our code is available at \url{https://github.com/Yaoming95/CIAT}~.
\end{abstract}

% Multilingual machine translation has attracted attention due to its efficiency in translating across various languages by a universal model. 
% However, this approach usually yields inferior performance to the bilingual translation models, especially for rich-resource language pairs. 
% This paper presents a novel perspective to attribute such a performance gap to meaning conflation deficiency for multilingual word representation and multilingual confusion induced by model incapacity. 

\section{Introduction}
\label{sec:intro}

Machine translation~(MT) is a core task in natural language processing. 
In recent years, neural machine translation~(NMT) approaches have made tremendous progress and takes the lead in the  field~\cite{DBLP:journals/corr/BahdanauCB14,vaswani2017attention,johnson2017google}. 
Conventionally, each NMT model only tackles a single language direction (\eg\ English $\rightarrow$ German).
A commonly used model like Transformer has $S=240$ million parameters. 
Therefore, Translating $N$ language pairs requires training models separately for each direction, resulting in $S\times N$ total parameters.
%(assuming each model has $S$ parameters). 
The huge size of all models for every language direction can be too costly to deploy, considering more than 100 popular languages worldwide. 
Hence, developing a single unified yet parameter-efficient model for
multilingual machine translation, enabling the translation of multiple directions, becomes crucially important.
There is much effort towards multilingual machine translation.
\citet{johnson2017google} first proposed training a single multilingual model via an additional target language tag, which became the paradigm for training multilingual MT models henceforth~\citep{gu2018universal,tan2018multilingual,tan2019multilingual,DBLP:conf/acl/SiddhantBCFCKAW20}.
Their approach is simple and parameter-efficient,
however, it often lags behind separately trained bilingual models, especially on resource-rich language pairs, where the phenomenon is known as performance degradation~\citep{aharoni2019massively}.

This paper analyzes the performance degradation between multilingual MT models and separate bilingual ones. 
Prior research suggests that unified embedding on a joint vocabulary leads to meaning conflation deficiency in multilingual training~\citep{camacho2018word}. 
For example, words with identical spelling may have distinct meanings in different languages --- \emph{bride} in English refers to a woman soon to get married while in French it means horse bridle. 
When it comes to machine translation, the effect goes beyond word embedding.
As a single model has bounded capacity, the multilingual learning may cause negative influences among shared parameters~\cite{liu2017adversarial, zhang-etal-2020-improving}.
We conjecture that such a performance degradation is due to the interference across languages brought by joint training on multiple language directions. 
Such interference affects both joint token embedding and representations from intermediate layers. 
We argue that resolving the interference is critical to improving multilingual translation performance.

Inspired by the insights above, we propose \emph{Counter-Interference Adapter} for multilingual machine Translation~(\mymethod).
The \mymethod includes a major multilingual base model (i.e., a multilingual Transformer), which is universally pre-trained on multiple language directions, and two kinds of designed adapter modules, which are trained on specific language directions. 
Specifically, we propose \textit{embedding adapter} and \textit{layer adapter} to reduce multilingual interference among embedding and intermediate layers, respectively. We also seek a new parallel connection method for adapter units, which is more effective for multilingual MT than the previous series connection. We validate \mymethod among three real-world datasets and observe that \mymethod gains significant improvements in translation performance over all other multilingual baselines.

Our contributions are as follows: 
\begin{inparaenum}[\it a)]
\item  We analyze performance degradation in multilingual NMT and formulate it as two issues; 
\item We propose \mymethod, an adapter-based framework to tackle the two issues and enhance the performance of the multilingual model with small amounts of extra parameters; 
\item We demonstrate the efficacy of \mymethod through extensive experiments on IWSLT, OPUS-100, and WMT benchmark datasets, surpassing other multilingual models over most of the translation directions.
\end{inparaenum}

\section{Related Work}

\paragraph{Multilingual Machine Translation.}
The multilingual MT enjoys a rich research history, dating back to the age of statistical machine translation~\cite{gao2002mars,haffari2009active,seraj2015improving}. In recent years, the prosperity of neural machine translation~(NMT) has led to the growing prominence and popularity of multilingual MT systems. The encoder-decoder framework has made the de facto standard for NMT~\cite{DBLP:journals/corr/BahdanauCB14,vaswani2017attention}. 
\citet{dong2015multi} did the pioneering work on extending conventional NMT to one-to-many translation, where the authors added a distinct decoder for each target language. 
\citet{firat2016multi} further extended such framework into many-to-many settings by building exclusive encoders and decoders for each language. Those attempts still faced problems such as low parameter utilization. 
%, since each language was still modeled independently explicitly. 
On the other hand, \citet{lee2017fully} treated all sources as the same language by translating on a character level. However, it only meets the many-to-one scenarios. 
\citet{johnson2017google} managed to train a single model that applied to multiple translation directions. Their solution is relatively simple: they attached a dedicated token at the beginning of the source sentence to specify the target language, while the rest of the model was shared among all languages. The paper has set a milestone of multilingual MT and has become the basis for most subsequent work. 

Recent studies paid more attention to the performance improvement of multilingual models based on \citet{johnson2017google}'s effort.
Several improved the model with external knowledge from human or other models:
\citet{tan2018multilingual} boosted the multilingual model by knowledge distillation,
\citet{tan2019multilingual} pre-clustered languages to assist similar languages. 
Several studies enhance the model from data: 
%Balancing training for multilingual neural machine translation
\citet{xia2019generalized} and \citet{DBLP:conf/acl/SiddhantBCFCKAW20} conducted data augmentation to low-resource languages via related high-resources or monolingual data.
\citet{taitelbaum2019multilingual} improved translation with relevant auxiliary languages.
Some other studies enhanced the Transformer model by introducing language-aware modules and learning language-specific representation~\cite{wang2019compact,zhu2020language}.

\paragraph{Adapter Network for Machine Translation.}
Our design derives from the residual adapters of the domain adaptation task. 
Concretely,~\citet{rebuffi2017learning} proposed the residual adapters in the computer vision area. They appended small networks (named adapters) to a pre-trained base network and only tuned the adapter on the specific task. \citet{houlsby2019parameter} adopted the idea into NLP domain adaptation tasks and designed the adapter for the Transformer, as shown in Fig.~\ref{fig:SeriesAdapter}. \citet{bapna2019simple} further extended the model to MT domain adaptation, and they regarded multilingual MT as a domain adaptation task. Based on their design, \citet{philip2020language} proposed the monolingual adapter for easy extension to new pairs, and \citet{zhang2021share} introduced conditional language-specific routing strategy(CLSR) to enhance model capacity in language-specific representation. 
However, their adapter designs followed a \textit{serial} connection manner, which might be limited for multilingual MT. We will discuss this in the following sections. 

%\cite{houlsby2019parameter} and~\cite{bapna2019simple} acclimated the idea into NLP domain adaptations by add a similar structure to every layer of Transformer\cite{vaswani2017attention}. We conclude their design in Fig~\ref{fig:SeriesAdapter}. Generally, a small neural network with a residual connection is attached to the tail of each layer of the Transformer. The attached adapter transforms and modifies $i$-th layer's output $x_{i+1}$,  rather than adapts its input $x_{i}$. Slightly different from~\cite{rebuffi2017learning},~\cite{bapna2019simple} only trained the parameters of the standard Transformer in global pre-training, and non-trained adapter layers are injected instantly during domain-specific adaptation.

%where they append a simple $1 \times 1$ filter bank with residual connection after each layer of the main convolutional neural networks. The whole network is pre-trained on the original task, thereafter only the adapters are fine-tuned on the specific task.

%Our work is closest to the last one above, who introduced residual adapter units to two kinds of machine translation task: domain adaptation and multilingual translation. We point out the potential limitations of their design in the next following sections and offer our solution.

\section{Challenges on Multilingual Machine Translation}
\label{sec:Preliminaries}
% \yaoming{直接叫Challenges会不会太泛？}

The ultimate goal of multilingual machine translation is to build a universal model to achieve mutual translation among all natural languages. That is, given a source sentence $\mathrm{s}$ and the target language $l$, the multilingual MT system shall output a sentence that resembles human reference $\mathrm{t}$.
% Under neural MT configurations, the model tries to learn the parameters $\theta$ which maximizing the likelihood $\log P(\mathrm{t}|\mathrm{s}, l; \theta)$.

Currently, Transformer~\cite{vaswani2017attention} gains popularity and becomes the paradigm for state-of-the-art NMT systems. 
Here, we follow the recent implementations~\cite{klein2017opennmt,vaswani2018tensor2tensor} of the pre-norm transformer, whose layer normalization is applied to the input of each sub-layer. The transformation of $i$-th sub-layer taking $x_i$ as input can be formulated as:
\mathchardef\mhyphen="2D
 \begin{equation}
    x_{i+1} = F_{\theta} (x_i) =  \mathrm{sub\mhyphen layer}_{\theta} (\text{LN}\left(x_i\right)) + x_i
\label{Eq:basic}
\end{equation}
where $\text{LN}\left(\cdot \right)$ is the layer normalization function and  
$\mathrm{sub\mhyphen layer}$ function denotes one basic layer, that is, self-attention, cross-attention or feed-forward layer. 
$\theta$ is the trainable parameters of the sub-layer.

The classic multilingual translation approach~\cite{johnson2017google} takes sentences from all language pairs and results in a universal model that translates across various languages. 
However, as previous researches have discovered, the multilingual model yields an inferior performance on high-resource languages compared to the bilingual models under the same configuration. 
In this paper, we reconsider the limitations of multilingual models and attribute the performance degradation to the following two issues:

% \noindent \textbf{Multilingual Embedding Deficiency}
\paragraph{Multilingual Embedding Deficiency}
% Proper word embedding boosts the performance of downstream tasks.
\citet{camacho2018word} addressed the \textit{meaning conflation deficiency} problem of the word embedding as a single vector is limited for representing polysemy. We extend the meaning conflation deficiency into the multilingual scenario. Generally, words/tokens may have unrelated or even opposite meanings in different languages. For example, ``娘" \ denotes mother in Chinese but daughter in Japanese. As the word embeddings are usually jointly trained on a multilingual corpus, representing a multilingual word with just one single vector may burden the model's semantic representation. 
We refer to the problem as \textit{multilingual embedding deficiency}. 

% In multilingual models, word embeddings are usually jointly learned, which may lead to \textit{meaning conflation deficiency}~\cite{camacho2018word} as many words enjoy multi-sense in a multilingual environment. 

% \noindent \textbf{Multilingual Interference Effects}
\paragraph{Multilingual Interference Effects}
Besides the word embedding, the insufficient capacity of a single NMT model also bottlenecks its performance on multilingual tasks ~\cite{aharoni2019massively,zhang-etal-2020-improving}.
The parameter-sharing among different languages may be a potential cause of negative interference~\cite{liu2017adversarial,wang2020negative}. 
We here formulate this phenomenon as \textit{multilingual interference effects}; that is, when a single model tries to learn multiple languages simultaneously, the extracted language features interfere with each other impose adverse effects upon overall performance.
Accordingly, we regard the model trained on bilingual data offers the approximately optimal solution on language representation, compared to which the representation of the multilingual model is bias-influenced:
\begin{equation}
\label{eq:m_eq_b_add_n}
    F_{\theta^m}(x_i) = F_{\theta^b}(x_i)  + \delta_i 
\end{equation}
%= x_{i+1} - \delta_i
where $\theta^{b}$ denotes the parameters of bilingual baselines and $\theta^{m}$ indicates the multilingual model. $\delta_i$ is the interference noise in the $i$-th layer.

\section{Proposed Method}

\begin{figure}[!t]
\centering
\includegraphics[width=.99\linewidth]{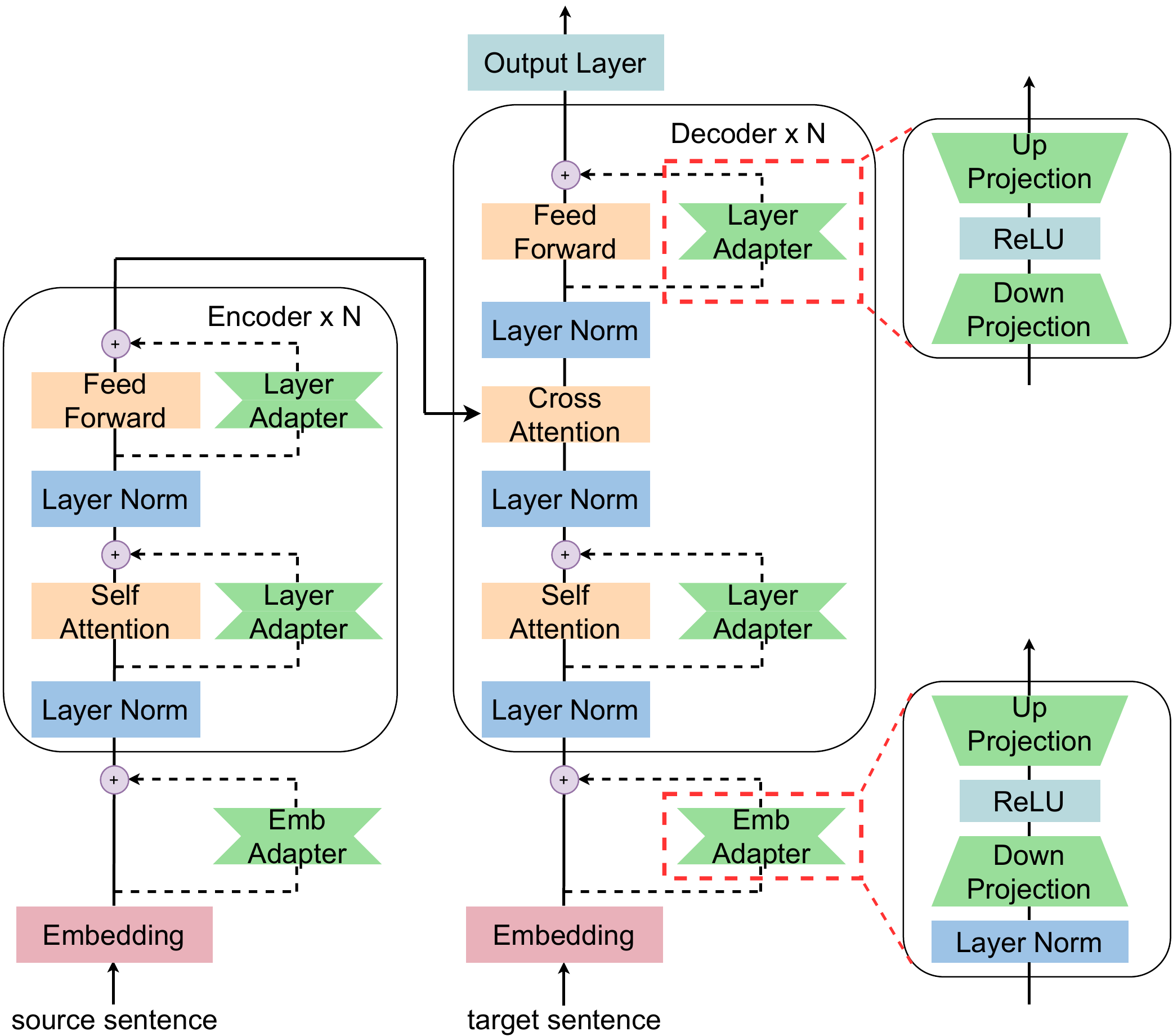}
\caption{The schematic diagram shows the overall layout of the model and the architecture of the adapters. First, the standard Transformer (modules connected by the solid line) is trained on the entire corpus as a multilingual base model, and its parameters are frozen afterward. The adapters (modules connected by the dashed line) are then plugged into the base model and fine-tuned on the bilingual corpus to enhance model performance on the specific language pair.}
\label{fig:layout}
\end{figure}

We propose a Counter-interference Adapter for Multilingual Machine Translation~(\mymethod) to address the two issues mentioned above with adapter-based architectures: embedding adapter and layer adapter. 
Figure~\ref{fig:layout} illustrates the overall architecture of \mymethod, which we will describe in detail.

\subsection{Embedding Adapter}

As discussed in section \ref{sec:Preliminaries}, a jointly trained multilingual word embedding $\mathbf{E}^m$ could be problematic as a word may have different meanings among multiple languages.
Empirically, we can fine-tune the whole embedding matrix for each language pair to address the multilingual embedding deficiency. 
However, tuning the whole matrix is quite expensive as the embedding matrix occupies a large part of model parameters, which also violets the advantage of parameter sharing in multilingual NMT. 
We hence introduce the embedding adapter  to approximate the fine-tuned embedding matrix $\mathbf{E}^f$ with much fewer parameters:
\begin{equation}
\label{eq:emb-adapter}
\tilde{\mathbf{E}}^f [w] = \mathbf{E}^m[w] - G_{\psi}(\mathbf{E}^m[w])
\end{equation}
where $\mathbf{E}^{\cdot} [w]$ is the embedding vector of token $w$ under embedding $\mathbf{E}^{\cdot}$, and $\tilde{\mathbf{E}}^f$ is the approximation of the fine-tuned embedding matrix $\mathbf{E}^f$. $G_\psi$ is our proposed embedding adapter parameterized by $\psi$.

Fig.~\ref{fig:layout} shows the layout of the embedding adapter, including a layer normalization~\cite{ba2016layer} and a fully connected feed-forward neural network. Following the suggestions of \citet{houlsby2019parameter}, we choose \textit{bottle-neck} architecture for the adapter module to save the parameters: the first layer down-projects the embedding dimension $d$ to a smaller size $m$, while the second layer projects it back to $d$ dimension. We select ReLU~\cite{nair2010rectified} as the activation function for the middle layer while using no activation for the output layer.

% Based on the embedding adapter, we can simulate the fine-tuned embedding matrix $\mathbf{E}^f$ with much fewer parameters.

\subsection{Parallel De-noise Layer Adapter}

\begin{figure}[!tbp]
\centering
\begin{subfigure}[t]{.52\linewidth}
  \centering
  \includegraphics[height=1\linewidth]{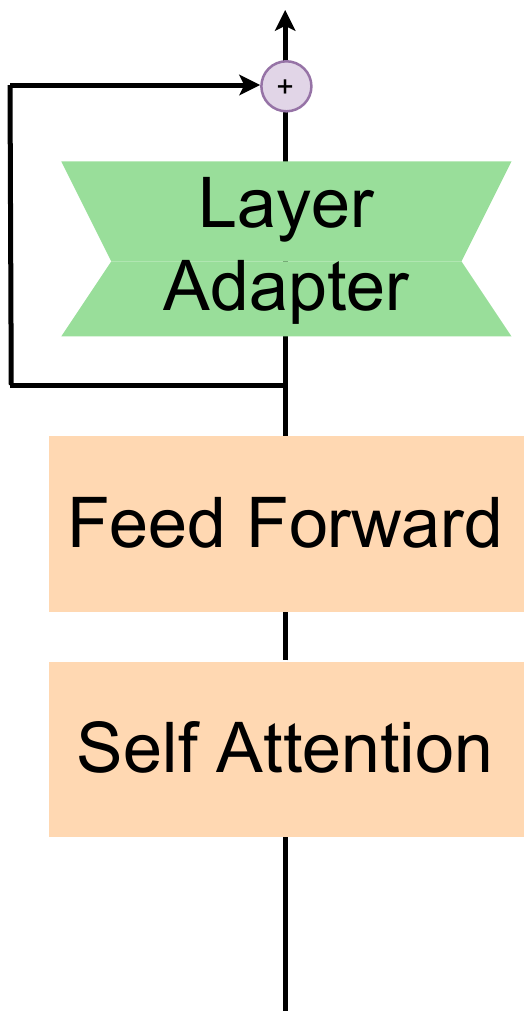}
  \caption{Serial}
  \label{fig:SeriesAdapter}
\end{subfigure}%
\begin{subfigure}[t]{.48\linewidth}
  \centering
  \includegraphics[height=0.8\linewidth]{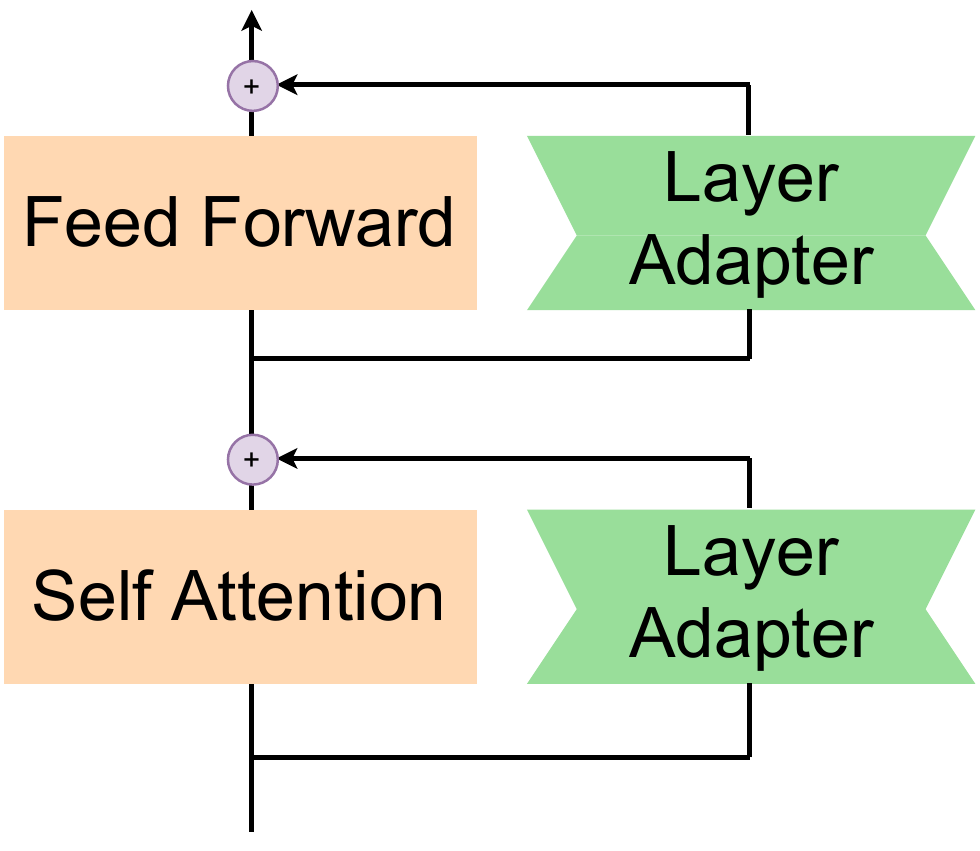}
  \caption{Parallel}
  \label{fig:ParallelAdapter}
\end{subfigure}
\caption{Architecture of Transformer with serial \cite{bapna2019simple} and parallel layer adapters~(ours), where the add-and-norm is omitted for simplicity. Compared to the Serial design, our parallel layer adapter design de-noise the multilingual interference \textit{pre} to the Transformer layers rather than \textit{post} to the layers.}
% do not include any extra residual connections and
\label{fig:test}
\end{figure}

As formulated in Eq.~\ref{eq:m_eq_b_add_n}, the multilingual representation $F_{\theta^m}(x_i)$ is regraded as a bias-influenced one compared to the bilingual representation $F_{\theta^b}(x_i)$.
To alleviate the interference, we introduce the layer adapter $G_\phi$ to model the bias term $\delta_i$:
% We formulate the multilingual interference effects in  Eq.~\ref{eq:m_eq_b_add_n}, where the multilingual representation $F_{\theta^m}(x_i)$ is regraded as a bias-influenced one compared to the bilingual representation $F_{\theta^b}(x_i)$.
% We hence hope to alleviate the bias term $\delta_i$ in Eq.~\ref{eq:m_eq_b_add_n}. Naturally, we can offset the interference by directly introducing a new module $G_\phi$ to approximate the opposite of $\delta_i$: 
\begin{equation}
\label{eq:para}
    G_{\phi}(x_i) \approx -\delta_i = F_{\theta^b}(x_i) - F_{\theta^m}(x_i)
\end{equation}
Eq.~\ref{eq:para} means that the layer adapter shares the same input $x_i$ as the sub-layer and de-noises the output $F_{\theta^m}(x_i)$. As a result, the output of $i$-th sub-layer is adapted to $x^\prime_{i+1} = F_{\theta^m}(x_i)+G_{\phi}(x_i)$.  

We connect layer adapters parallel to sub-layers, as shown in Fig.~\ref{Tab:overall}. The architecture of the layer adapter is the same as the embedding adapter, except for the removed layer normalization. In practice, we find that the layer adapter can share the layer normalization structure with the corresponding sub-layer to achieve the best performance.
%

%Note that we regard the sub-layers (self-attention or feedforward network) as the basic unit of the Transformer in our design. 
%As the Transformer and the adapters are parallelly combined, the adapter receives the original input $x_i$ other than the biasd  $F_{\theta^m}(x_i)$.
% \begin{equation}
%     G_{\phi}(x_i) \sim -\delta_i = F_{\theta^b}(x_i) - F_{\theta^m}(x_i)
% \end{equation}
%In our design, the adapters can be thought $G$ as a debias function, learned to estimate and reduce the introduced bias of each sub-layer $F$. 

\paragraph{Comparison to the Serial Adapter Design}

As the first study introduced adapter networks to machine translation,~\citet{bapna2019simple} also conducted experiments on multilingual machine translation.
They append adapters \textit{serial} to the model architecture with a residual connection as illustrated in Fig.~\ref{fig:SeriesAdapter}, while we argue that our \textit{parallel} connection is more suitable for multilingual machine translation.
Compared to Eq.~\ref{eq:para}, we can formulate the serial style adapters as:
\begin{equation}
\label{eq:seriesAdapter}
 G_\phi (F_{\theta^m}(x_i) ) \approx  - \delta_i= F_{\theta^b}(x_i) - F_{\theta^m}(x_i)
\end{equation}
where the adapter $G$ receives the bias-influenced hidden states $F_{\theta^m}(x_i)$ other than the original input $x_i$. However, the bias-influenced $F_{\theta^m}(x_i)$ may not be distinguishable for training adapters, which is especially the case when the multilingual model is inferior, making $F_{\theta^m}(x_i)$ fail to capture enough information of languages.

% As discussed in Eq.~\ref{eq:m_eq_b_add_n}, given the multilingual and bilingual models, the interference bias  $\delta_i=F_{\theta^m}(x_i) -  F_{\theta^b}(x_i) $ is determined by the sub-layer input $x_i$.
% ~\citet{bapna2019simple}'s design de-noise the multilingual interference $ \delta_i$ \textit{post} to the Transformer layer
% % learn to de-bias based on the output of the layer: 
% \begin{equation}
% \label{eq:seriesAdapter}
%  G_\phi (F_{\theta^m}(x_i) ) \approx  - \delta_i= F_{\theta^b}(x_i) - F_{\theta^m}(x_i)
% \end{equation}
% In their design, adapter $G$ receives the bias-influenced hidden states $F_{\theta^m}(x)$ other than the original input $x_i$. 
% However, the bias-influenced $F_{\theta^m}(x)$ might not be distinguishable for the training of adapters, which is especially the case when the multilingual model is inferior, making $F_{\theta^m}(x)$ fail to capture enough information of languages(\eg $F_{\theta^m}(x)$ is almost the same for two distinct pairs under the interference). 

In contrast, our parallel design de-noise the bias-influenced term \textit{pre} to the sub-layers. Corresponding to Eq.~\ref{eq:para}, the parallel layer adapters receive the same input as the sub-layers and de-noise directly to the output, which is a more intuitive and natural design for de-noising $\delta_i$, since the adapter is independent of the sub-layer output. The parallel adapter can also be regarded as the low-rank ``patch'' for the corresponding sub-layer, which adjusts the parametrization of the high-rank sub-layer to the specific language pair and fix the multilingual interference.

% We let the layer adapters share the layer normalization of its corresponding Transformer layer, while each embedding adapter has its layer normalization exclusively. 
% Compared to the serial design of \citet{bapna2019simple}, our layer adapters do not include any extra residual connection or layer normalization, leading to a simpler network and fewer parameters for each adapter unit. 

\subsection{Model Training}
The training process of the whole model consists of two phases: the pre-training on the multilingual model and the learning of the adapter modules.
First, we pre-train the standard Transformer on the entire corpus, making a universal multilingual model.
The parameters of the Transformer are frozen once the model converges.
Then, we ``plug-in'' the randomly initialized adapters for each specific language pair and only fine-tune the adapter parameters on this pair. 
Since the base Transformer model is frozen, each plugged adapter's learning process is independent of other adapters.
During the inference stage, we only apply the base model and the corresponding adapter to translate sentences into the target language.
\footnote{Note that \citet{bapna2019simple} only fine-tune their model on high-resource pairs, while we find \mymethod can be applied to both high-resource and low-resource pairs.} All the adapters are plug-able during the inference stage: disabling the adapter will degenerate the model into a basic multilingual translation model, and when the model is required to translate a specific pair, we just ``plug-in'' the specific adapter into the base model.
%while enabling the certain adapter will transfer the base model to translate the specific language pair.

Theoretically, the well-trained adapted network guarantees a better performance compared to the multilingual base. When the layer adapter is disabled (\ie. $G_{\phi}(x_i)\equiv0$ when all parameters are zero), the model is reduced to the multilingual base model. With proper training, the layer adapter should boost the model performance. 

\section{Experiments}
\label{Sec:Exp}
We conducted experiments on three multilingual translation datasets to show the effectiveness of \mymethod. 

\subsection{Datasets}
We focused on two mainstream multilingual cases: many-to-English and English-to-many, since the many-to-many case can be bridged via English as a pivot.
We collected the following three datasets for our experiments: 

\noindent\textbf{IWSLT}~\footnote{\url{https://wit3.fbk.eu}} is a small dataset from TED talks, where we used 8 languages $\leftrightarrow$ English from year 2014 to 2016 release. 

\noindent\textbf{OPUS-100}~\cite{zhang-etal-2020-improving}~\footnote{\url{http://opus.nlpl.eu/OPUS-100.php}} is an English-centric dataset covering 99 languages $\leftrightarrow$ English pairs. We selected 20 language pairs, 17 of which have 1 million data samples while 3 language pairs are under low resource setting.%: be(67k), nb(142k) and af(275k). 

\noindent\textbf{WMT}~\cite{barrault2019findings} datasets are also involved, which contains five language pairs ranging from the year 2014 to 2019.

For simplicity, we use the ISO 639-1 code as the abbreviation for language names. The detailed data statistics are listed in the Appendix.
%and we release our scripts for collecting all above datasets at \url{https://anonymous.url}.
%~\footnote{\url{http://xml.coverpages.org/iso639a.html}}

\subsection{Implementation Details}
For each dataset, we tokenize sentences using SentencePiece~\cite{kudo2018sentencepiece} jointly learned on the source and target side, and we set vocabulary size to 32,000.
%~\footnote{\url{https://github.com/google/sentencepiece}}
For model setup, we follow the same configuration as~\citet{tan2018multilingual} on IWSLT, including 2 layers for both encoder and decoder. The embedding dimension was 256, and the size of feed-forward hidden units was 1,024. The attention head was set to 4 for both self-attention and cross-attention. 
For OPUS-100 and WMT, we follow the standard Transformer-Big setting~\cite{vaswani2017attention}, including 6 layers for encoder and decoder. The embedding dimension, feed-forward hidden size, and attention head were set to 1024, 4096, and 16, respectively. 
The hidden state's dimension of the adapters' inner layer are set to be half of the embedding size,~\ie~128 for IWSLT and 512 for OPUS-100 and WMT. 
We use Adam optimizer~\cite{kingma2014adam} with the same schedule algorithm as \citet{vaswani2017attention}. During Inference, we use a beam width of 4 and length penalty of 0.6.
%The base multilingual model and the adapters are trained with a global batch size of {\color{red}32,768?} and  {\color{red}32,768?} respectively on NVIDIA Tesla V100 GPUs. 

% \noindent \textbf{\mymethod-basic}: The model keeps only the parallel adapters for feed-forward networks so that the number of parameters is close to that of \citet{bapna2019simple}, which serves as an important competitor.

% \noindent \textbf{\mymethod-block}: The model keeps all layer adapters except for the embedding adapter to validate the embedding adapter's effectiveness.

All our experiments are evaluated by tokenized BLEU~\cite{papineni2002bleu} using \texttt{multi-bleu.perl}
\footnote{\url{https://github.com/moses-smt/mosesdecoder/blob/master/scripts/generic/multi-bleu.perl}}.
We implement our models via TensorFlow~\cite{abadi2016tensorflow} and train models on NVIDIA Tesla V100 GPUs.

% \subsection{Compared Models}
% \subsubsection{baselines}
% We compared the proposed Transformer-Adapter framework with the following baselines in order to validate the effectiveness of our design. 

% \noindent \textbf{Bilingual.} We train the vanilla bilingual NMT with the same configuration as a strong baseline.

% \noindent \textbf{Multilingual.} We also compare our model to the multilingual one, \ie~the base model without any adapter introduced.

% \noindent \textbf{Serial Adapter(Serial).} ~\cite{bapna2019simple} made the first attempt on applying adapter to machine translation. 

% \noindent \textbf{Knowledge Distillation(KD).}~\cite{tan2018multilingual} tackles multilingual machine translation with knowledge distillation, where bilingual models are first trained as the teachers, then the multilingual model are trained as student. 

\subsection{Main Results}

We compared our model with several strong baselines and effective models:

\noindent \textbf{Bilingual}: The model is trained with only bilingual data with the same model configuration, which serves as the strong benchmark.

\noindent \textbf{Multilingual}~\cite{johnson2017google}: The data of all language pairs are mixed to train the model.

\noindent \textbf{Knowledge Distillation~(KD)}~\cite{tan2018multilingual}~\footnote{\url{https://github.com/RayeRen/multilingual-kd-pytorch}. Note that we remove the \textit{lowercase} option in their preprocessing script.}: The bilingual models are first trained as the teachers, then the multilingual models are trained as students.
%We reproduce the results using their released code.
We only conduct KD on IWSLT due to computational resource limitations. 

\noindent  \textbf{Serial}~\cite{bapna2019simple}: We re-implement the series adapter model as illustrated in Fig~\ref{fig:SeriesAdapter}, which made the first attempt on applying adapters to machine translation. 

\begin{table}[t]
\centering
\resizebox{1\columnwidth}{!}
{
\begin{tabular}{c|cccccc}
\hline
                               & \multicolumn{2}{c|}{en-IWSLT} & \multicolumn{2}{c|}{en-OPUS-100} & \multicolumn{2}{c}{en-WMT}   \\ \hline
                               & $\leftarrow$  & $\rightarrow$ & $\leftarrow$   & $\rightarrow$  & $\leftarrow$ & $\rightarrow$ \\ \hline
Bilingual                      & 32.42         & 23.44         & 31.58          & 28.13          & 29.03        & 26.92         \\ \hline
Multilingual                   & 31.56         & 20.69         & 35.75          & 29.59          & 30.06        & 25.47         \\
KD                             & 30.97         & 20.31         & -              & -              & -            & -             \\
Serial                         & 31.63         & 21.37         & 36.44          & 31.26          & 30.69        & 26.92         \\ \hline
\mymethod-basic & 31.75         & 22.06         & 36.78          & 31.96          & 31.12        & 27.34         \\
\mymethod-block & 32.31         & 22.13         & 36.89          & 32.05          & 31.14        & \textbf{27.66}         \\
\mymethod       & \textbf{32.39}         & \textbf{22.48}         & \textbf{36.91}          & \textbf{32.07}          & \textbf{31.24}        & 27.63         \\ \hline
WR\%       & 87.5         & 100         & 25          & 75          & 60        & 80         \\ \hline
\end{tabular}
}
\caption{Overall Performance. The overall score is the arithmetic mean of the case-sensitive tokenized BLEU score of the test set of all languages. WR\% is the win ratio(\%), which denotes  the percentage of the language pairs in which \mymethod exceeds all multilingual baselines by \textit{at least} 0.5 BLEU.
}\label{Tab:overall}
\end{table}

%\mymethod+emb     & 16.06M   & 31.53        & 14.37         & 34.59        & 25.26         & 40.50        & 35.05         & 23.71        & 12.72         & 35.78        & 24.07         & 34.37        & 30.17         & 36.65        & 29.83         & 23.85        & 14.15   \\ \hline

\begin{table*}[!ht]
\centering
\scriptsize
\resizebox{2\columnwidth}{!}
{
\begin{tabular}{c|cccccccccccccc}
\hline
             & \multicolumn{2}{c|}{params(M)} & \multicolumn{2}{c}{en-ar}    & \multicolumn{2}{c}{en-fa}    & \multicolumn{2}{c}{en-de}    & \multicolumn{2}{c}{en-nl}    & \multicolumn{2}{c}{en-af}    & \multicolumn{2}{c}{en-da}    \\
             & \multicolumn{2}{c|}{}               & $\leftarrow$ & $\rightarrow$ & $\leftarrow$ & $\rightarrow$ & $\leftarrow$ & $\rightarrow$ & $\leftarrow$ & $\rightarrow$ & $\leftarrow$ & $\rightarrow$ & $\leftarrow$ & $\rightarrow$ \\ \hline
Bilingual    & \multicolumn{2}{c|}{242$\times N$}        & 37.45        & 23.11         & 22.10        & 9.92          & 33.22        & 30.21         & 30.24        & 26.96         & 46.63        & 44.79         & 35.80        & 35.32         \\
Multilingual & \multicolumn{2}{c|}{242}           & 39.94        & 22.92         & 24.34        & 10.29         & 34.55        & 29.67         & 33.73        & 28.59         & 53.31        & 45.67         & 38.12        & 36.21         \\
Serial       & \multicolumn{2}{c|}{242 $+$ 12.6$\times N$}          & 41.20        & 25.00         & 25.63        & 10.21         & 35.53        & 31.78         & 34.16        & 29.88         & 55.70        & 50.92         & 38.50        & 37.58         \\
\mymethod         & \multicolumn{2}{c|}{242 $+$ 27.3$\times N$}         & \textbf{41.58}        & \textbf{25.34}         &\textbf{ 25.77 }       & \textbf{11.03}         & \textbf{35.89}        & \textbf{32.76}         & \textbf{34.53 }       & \textbf{30.25 }        &\textbf{ 56.02  }      & \textbf{52.17}         & \textbf{39.38}        & \textbf{39.02}         \\ \hline
             & \multicolumn{2}{c}{en-no}           & \multicolumn{2}{c}{en-sv}    & \multicolumn{2}{c}{en-nb}    & \multicolumn{2}{c}{en-es}    & \multicolumn{2}{c}{en-pt}    & \multicolumn{2}{c}{en-fr}    & \multicolumn{2}{c}{en-it}    \\
             & $\leftarrow$     & $\rightarrow$    & $\leftarrow$ & $\rightarrow$ & $\leftarrow$ & $\rightarrow$ & $\leftarrow$ & $\rightarrow$ & $\leftarrow$ & $\rightarrow$ & $\leftarrow$ & $\rightarrow$ & $\leftarrow$ & $\rightarrow$ \\ \hline
Bilingual    & 25.20            & 27.31            & 30.63        & 30.56         & 32.36        & 29.93         & 36.78        & 36.52         & 34.82        & 31.56         & 33.83        & 35.11         & 33.28        & 30.54         \\
Multilingual & 27.92            & 29.19            & 32.88        & 33.03         & 43.48        & 35.52         & 41.01        & 37.60         & 38.37        & 33.03         & 35.33        & 34.20         & 37.22        & 31.57         \\
Serial       & 28.21            & 30.24            & 33.74        & 34.15         & 45.17        & 39.38         & 41.34        & 39.32         & 38.96        & 34.59         & 36.41        & 36.02         & 37.82        & 32.94         \\
\mymethod         & \textbf{28.43 }           & \textbf{31.67}            & \textbf{34.39 }       & \textbf{34.78}         & \textbf{45.82 }       & \textbf{40.46 }        & \textbf{41.73 }       & \textbf{40.12  }       & \textbf{39.16}        & \textbf{35.38}         & \textbf{36.49}        & \textbf{36.81}         & \textbf{37.88}        &\textbf{ 33.53}         \\ \hline
             & \multicolumn{2}{c}{en-pl}           & \multicolumn{2}{c}{en-cs}    & \multicolumn{2}{c}{en-sk}    & \multicolumn{2}{c}{en-ru}    & \multicolumn{2}{c}{en-uk}    & \multicolumn{2}{c}{en-be}    & \multicolumn{2}{c}{en-zh}    \\
             & $\leftarrow$     & $\rightarrow$    & $\leftarrow$ & $\rightarrow$ & $\leftarrow$ & $\rightarrow$ & $\leftarrow$ & $\rightarrow$ & $\leftarrow$ & $\rightarrow$ & $\leftarrow$ & $\rightarrow$ & $\leftarrow$ & $\rightarrow$ \\ \hline
Bilingual    & 24.39            & 20.56            & 31.63        & 25.22         & 36.84        & 30.58         & 32.75        & 29.30         & 24.26        & 16.15         & 7.72         & 7.71          & 41.74        & 41.29         \\
Multilingual & 28.16            & 21.94            & 36.46        & 26.51         & 40.02        & 31.48         & 35.207        & 29.07         & 28.97        & 16.43         & 25.22        & \textbf{19.85}         & 40.72        & 38.98         \\
Serial       & 28.63            & 23.78            & 36.51        & 28.59         & 41.19        & 33.26         & 35.78        & \textbf{31.05}         & 29.26        & 18.65         & 23.04        & 16.89         & 42.10        & 40.92         \\
\mymethod         & \textbf{28.88}            & \textbf{24.42 }           & \textbf{36.77}        &\textbf{ 28.70}         & \textbf{41.70}        & \textbf{33.89 }        &\textbf{ 35.93}        & 30.41         & \textbf{29.68  }      & \textbf{19.56 }        & \textbf{26.00 }       & 19.56         &\textbf{ 42.13}        & \textbf{41.54}         \\ \hline
\end{tabular}
}
\caption{Case-sensitive tokenized BLEU on the OPUS-100 dataset. params(M) represents the total number of parameters required for each model in million, where $N$ is the number of language pairs. }
\label{Tab:opus_main}
\end{table*}
%The results of two model variants are omitted here for simplicity, which can be referred to in the Appendix.

\begin{table*}[!h]
\centering
\scriptsize
\resizebox{1.8\columnwidth}{!}
{
\begin{tabular}{c|c|cccccccccc}
\hline
             & params(M) & \multicolumn{2}{c}{en-es}    & \multicolumn{2}{c}{en-de}    & \multicolumn{2}{c}{en-et}    & \multicolumn{2}{c}{en-ru}    & \multicolumn{2}{c}{en-lv}    \\
             &                & $\leftarrow$ & $\rightarrow$ & $\leftarrow$ & $\rightarrow$ & $\leftarrow$ & $\rightarrow$ & $\leftarrow$ & $\rightarrow$ & $\leftarrow$ & $\rightarrow$ \\ \hline
Bilingual    & 242 $\times N$           & 35.05        & 34.45         & 33.07        & 28.37         & 24.27        & 19.20          & 34.42        & 35.69         & 18.34        & 16.89         \\ \hline
Multilingual & 242           & 34.92        & 33.74         & 32.53        & 26.44         & 26.85        & 18.14         & 35.37        & 33.14         & 20.65        & 15.91         \\
Serial       & 242 $+$ 12.6 $\times N$        & 35.47        & 33.78         & 33.38        & 26.88         & 27.76        & 20.30          & 35.86        & 35.11         & 20.97        & 18.54         \\ \hline
\mymethod         & 242 $+$  27.3 $\times N$         & \textbf{35.75}         & \textbf{34.51} & \textbf{33.62 }         & \textbf{27.78}          & \textbf{28.33} & \textbf{20.97} & \textbf{36.9} & \textbf{35.97}          & \textbf{21.59} & \textbf{18.94}       \\ \hline
\end{tabular}
}
\caption{Case-sensitive tokenized BLEU on the WMT dataset. params(M) represents the total number of parameters required for each model in million, where $N$ is the number of language pairs. }
\label{Tab:wmt_main}
\end{table*}

To further analyze the impact of each component and compare the two designs of Serial and Parallel layer adapter under the same number of parameters, we introduce two model variants, named \mymethod-layer and \mymethod-block, and conduct the ablation study. \mymethod-layer has the same number of parameters compared with Serial and \mymethod-block removes embedding adapter. The ablation details are illustrated in the appendix. 

We also reproduced a recently proposed adapter-based MT model, namely Mono~\cite{philip2020language}. The model also follow a serial adapter connection manner, which we replace with a parallel connection to validate the effectiveness of our parallel design.
We show the comprehensive results of the ablation study and parallel variant of Mono in the Appendix. We here list the overall results of two model variants.

We present the BLEU score of three datasets on Table~\ref{Tab:opus_main},~ \ref{Tab:wmt_main} and \ref{Tab:iwslt_main}~(Appendix) respectively, and summarize the overall results on Table~\ref{Tab:overall}. 
Since the number of parameters is quite different among distinct baseline and model variants, we also indicate parameters amount for each model with respect to the number of translation directions. We present our findings as follows:

On the IWSLT dataset, the bilingual baselines are significantly better than the multilingual models, which align with our expectations. To our surprise, we observe the opposite phenomenon on OPUS-100, which may be because (1) the domain of sentences in the OPUS-100 dataset is close. (2) Most languages selected have relatives from the same language family, which leads to promotion between the related languages~\cite{tan2019multilingual}; while only \textit{zh} has no similar language in the OPUS-100, and bilingual performs better in \textit{zh}. 

Among various models, the gaps in BLEU scores of \textit{any}$\rightarrow$\textit{en} are smaller than that of \textit{en}$\rightarrow$\textit{any}. Meanwhile, \textit{any}$\rightarrow$\textit{en} direction gain less improvement from \mymethod and other baselines than \textit{en}$\rightarrow$\textit{any} direction. We attribute such phenomenon to the over-representation of English in the English-centric corpus~\cite{aharoni2019massively}, so it is more difficult to improve the generation quality of English than other languages.

\mymethod significantly outperforms other multilingual competitors among all datasets, and improve the BLEU score by at least 0.5 in 42 out of 64 language directions. On \textit{en}$\rightarrow$\textit{any} directions, the performance improvement is even more significant. 

We also find that the performance of \mymethod and baselines on different languages is also affected by language families and resource scarcity. According to Table~\ref{Tab:opus_main}, for languages of the same language family, \mymethod can better improve performance (e.g., Spanish, Portuguese, French, and Italian get a greater improvement than Arabic and Persian). And for low-resource languages(e.g. Afrikaans, Belarusian), compared with Serial baseline, \mymethod can bring more BLEU score improvement, especially when there are languages similar to these low-resources in the training set.

\subsection{Discussion on Adapters}

To obtain a comprehensive understanding of the adapters in \mymethod, we conduct a series of analyses on embedding adapters and layer adapters respectively. 
%, taking \textit{en}$\leftrightarrow$\textit{zh} directions from OPUS-100 as the study case.

\subsubsection{Semantic Alignment of Embedding Adapters}

To study whether the embedding adapter alleviates multilingual embedding conflation, we calculate the Average Cosine Similarity(ACS)~\cite{lin2020pre} of words with the same meaning across different languages to verify if embedding adapter help to align cross-lingual synonyms. We select top frequent 1000 words of five language pairs from MUSE bilingual dictionaries\footnote{\url{https://github.com/facebookresearch/MUSE}}, and compare the ACS results between vanilla multilingual embedding and ones with the \mymethod's embedding adapter, where the models are trained on OPUS-100 dataset. 

\begin{figure}[h]
\centering
\includegraphics[width=1\linewidth]{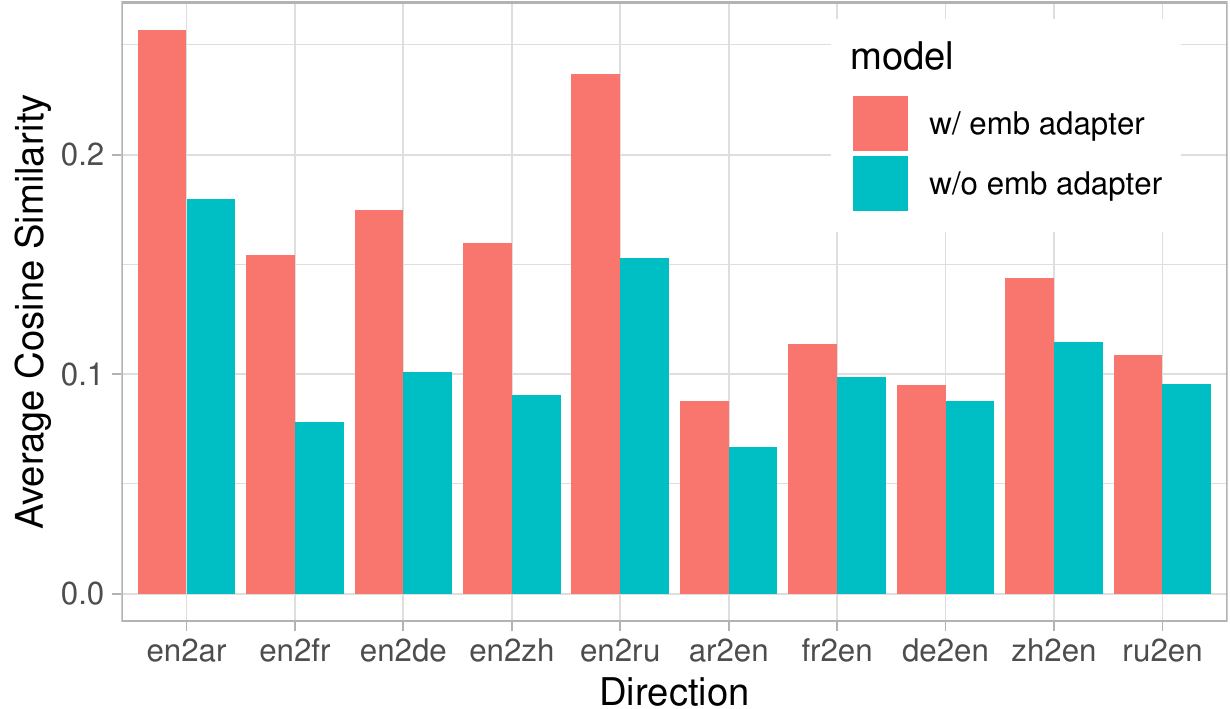}
\caption{Average cosine similarity between vanilla multilingual embedding (w/o emb adapter) and the adapted ones(w/ emb adapter). The ACS increases after applying the embedding adapter among all selected languages, indicating that embedding adapter reduce the embedding distance between cross-lingual synonyms.
}
\label{fig:ACS}
\end{figure}

We plot the ACS results in Fig.~\ref{fig:ACS} and observe two phenomena. Firstly, adding the embedding adapters increases ACS of the model among all selected pairs, which indicates that embedding adapters capture more semantic information between synonyms from different languages. The ACS improvement also suggests that embedding adapters indeed relieve the multilingual embedding conflation problem as the latent representation of synonyms between the two languages is brought closer with the auxiliary of the embedding adapter, which improves the overall performance. Secondly, compared to \textit{any} $\rightarrow$ \textit{en} directions, \textit{en} $\rightarrow$ \textit{any} pairs generally gain more improvement on ACS and BLEU scores, indicating embedding adapter is more effective on translating pivot language to other languages compared to the opposite direction.

\subsubsection{The Influence of Layer Adapters}

We also perform an extension experiment to discover the influence of layer adapters on the main model, taking \textit{en}$\leftrightarrow$\textit{zh} directions from OPUS-100 as the study cases.

\begin{figure}[!h]
\centering
\includegraphics[width=1\linewidth]{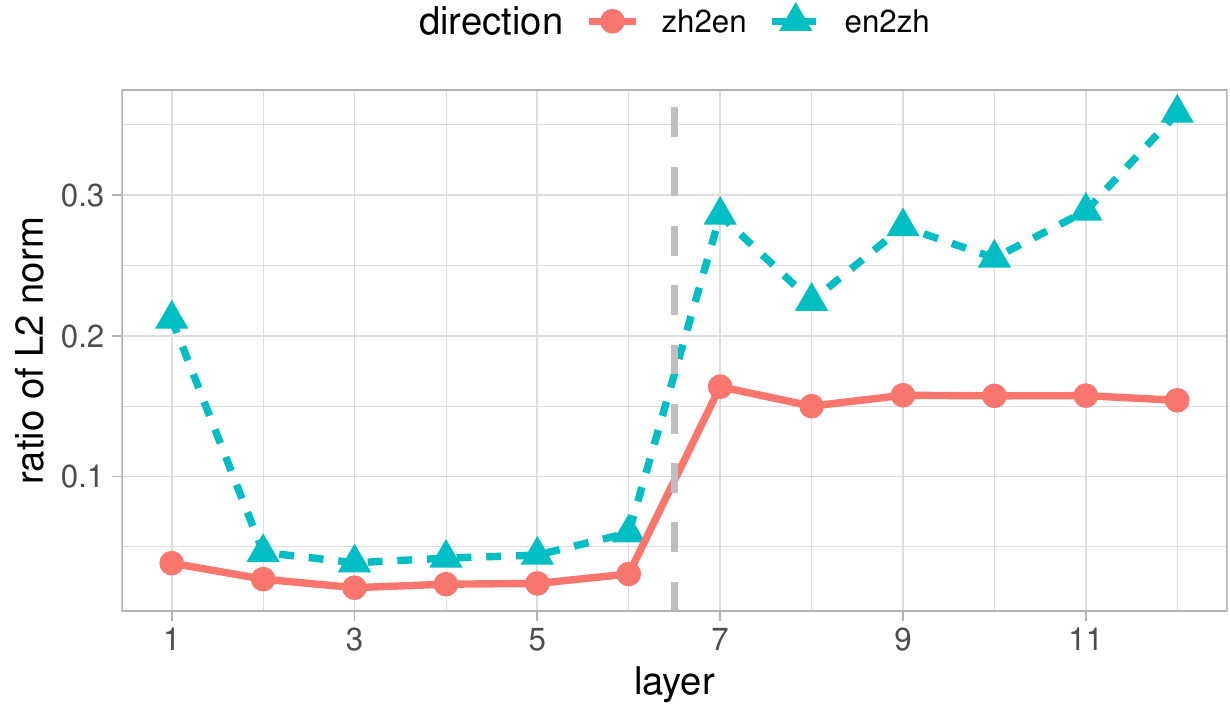}
\caption{
Hidden states L2-norm ratio of layer adapter over the base model in each layer. The first six layers are from the encoder, and the rests are from the decoder. The adapters of decoder influence more to the base model compared to the encoder ones. 
}
\label{fig:signal}
\end{figure}

We first plot the L2-norm ratio of the hidden state between the adapter and the base model across layers in Fig~\ref{fig:signal} to investigate how the ``plug-in'' adapter influence the multilingual model. In general, the adapters of decoder exerts a greater influence on the hidden states of the base model compared to the ones of the encoder, and the adapters provide a stronger signal in \textit{en}$\rightarrow$\textit{zh} compared to the opposite direction. 

\begin{figure}[!h]
    \centering
    \begin{subfigure}[b]{0.2\textwidth}
        \centering
        \includegraphics[width=\textwidth]{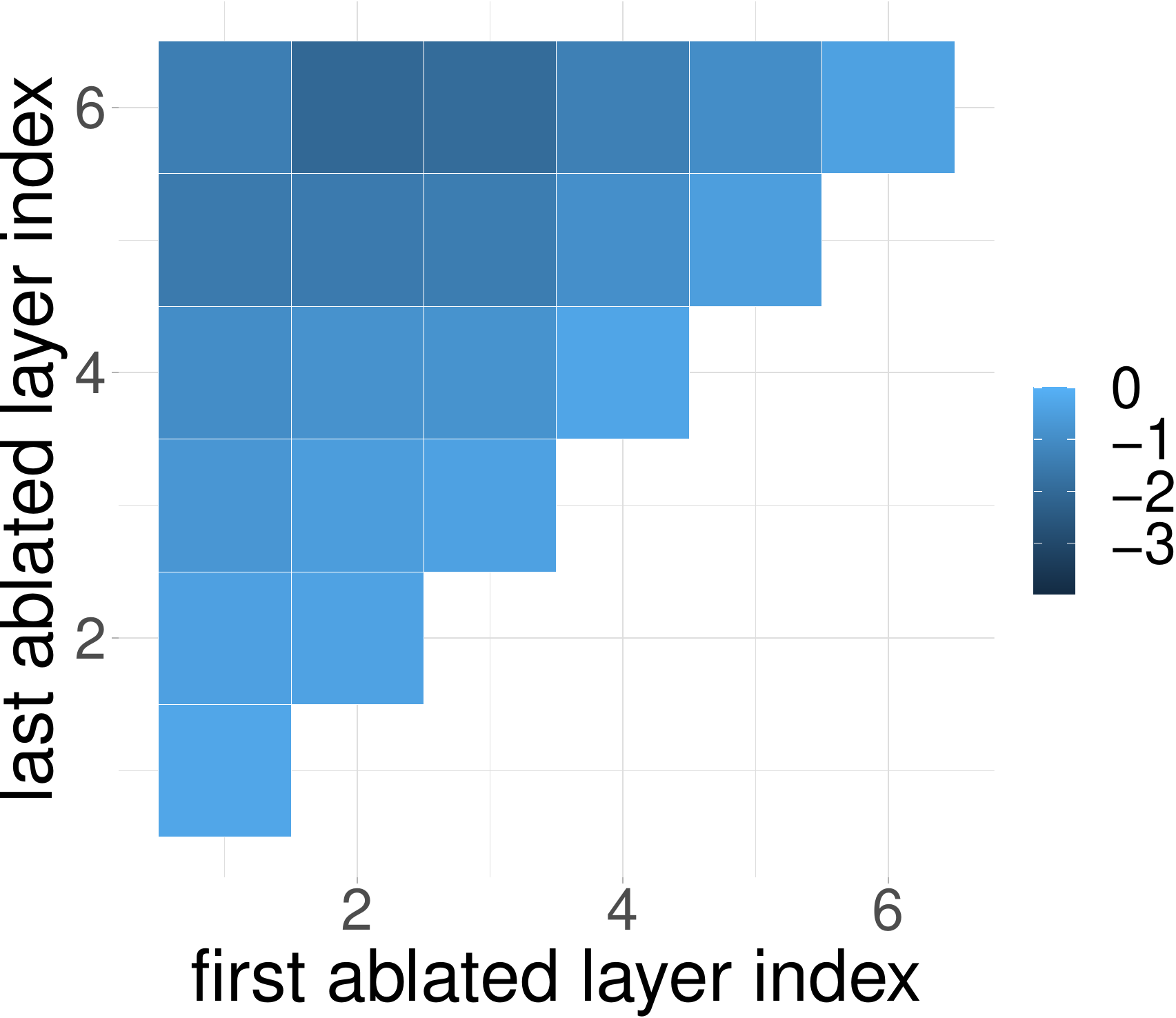}
        \caption[Network2]%
        {{\small encoder en2zh}}    
        %\label{fig:mean and std of net14}
    \end{subfigure}
    \hfill
    \begin{subfigure}[b]{0.2\textwidth}  
        \centering 
        \includegraphics[width=\textwidth]{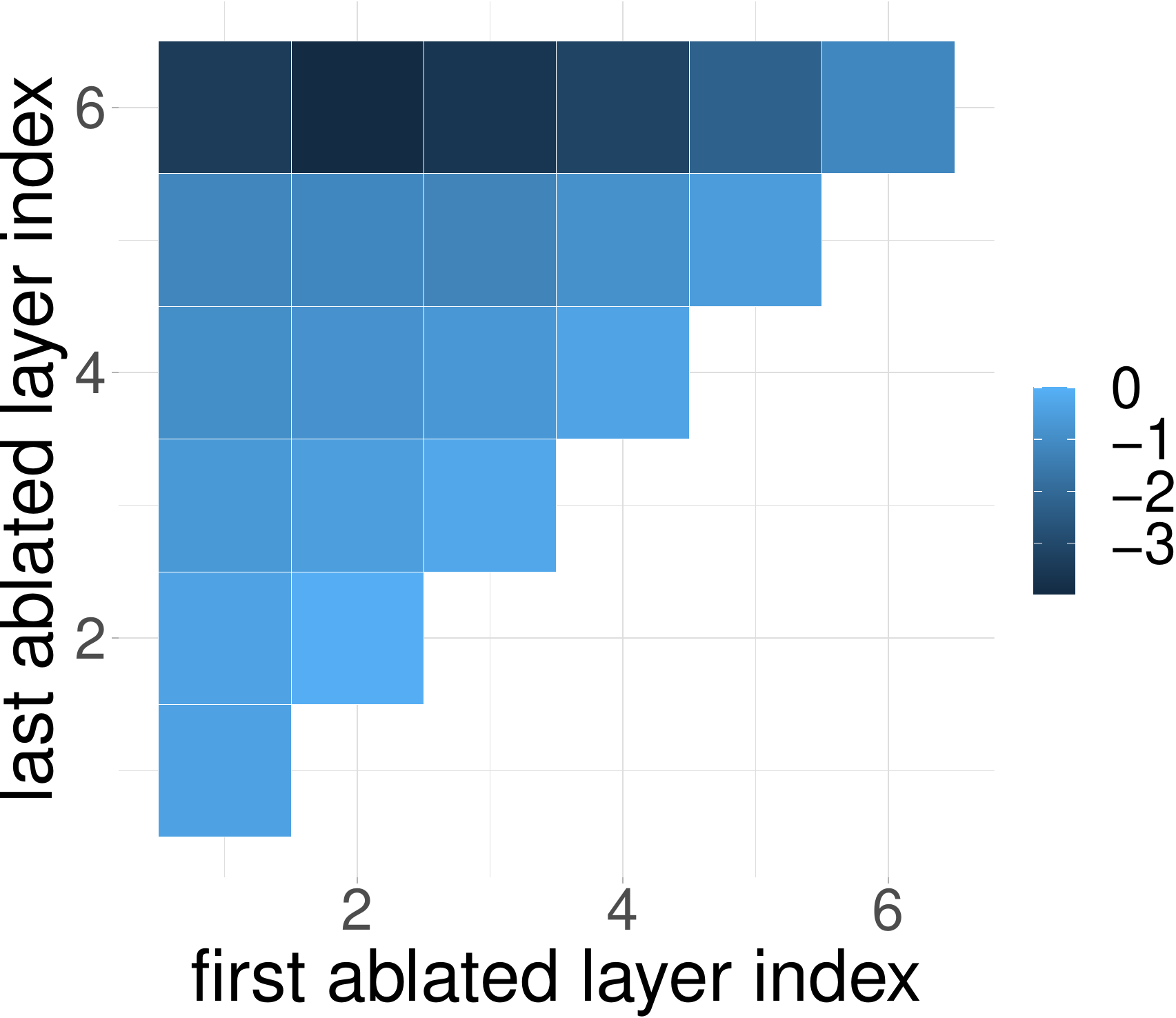}
        \caption[]%
        {{\small decoder en2zh}}    
        %\label{fig:mean and std of net24}
    \end{subfigure}
    \vskip\baselineskip
    \begin{subfigure}[b]{0.2\textwidth}   
        \centering 
        \includegraphics[width=\textwidth]{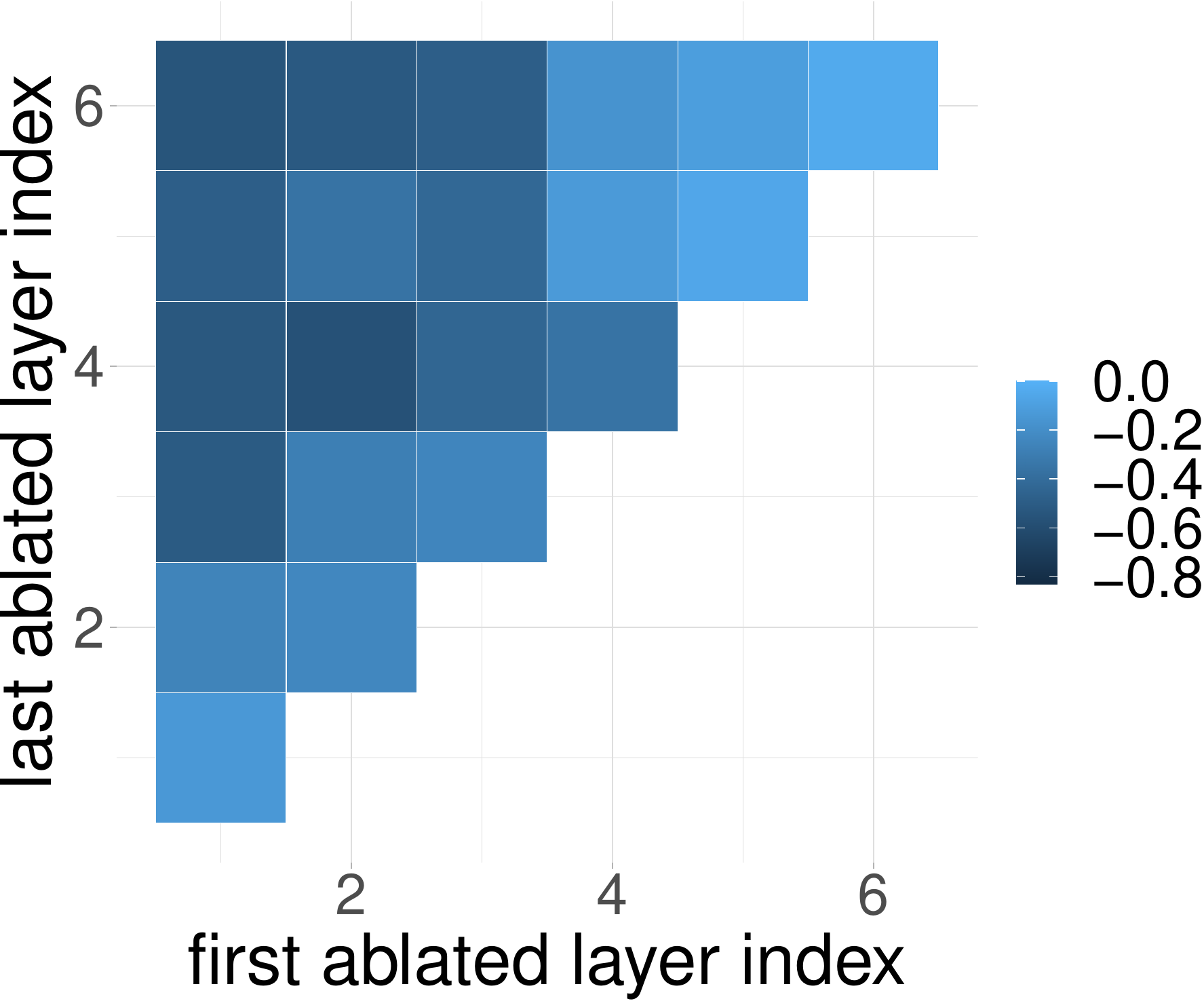}
        \caption[]%
        {{\small encoder zh2en}}    
        %\label{fig:mean and std of net34}
    \end{subfigure}
    \hfill
    \begin{subfigure}[b]{0.2\textwidth}   
        \centering 
        \includegraphics[width=\textwidth]{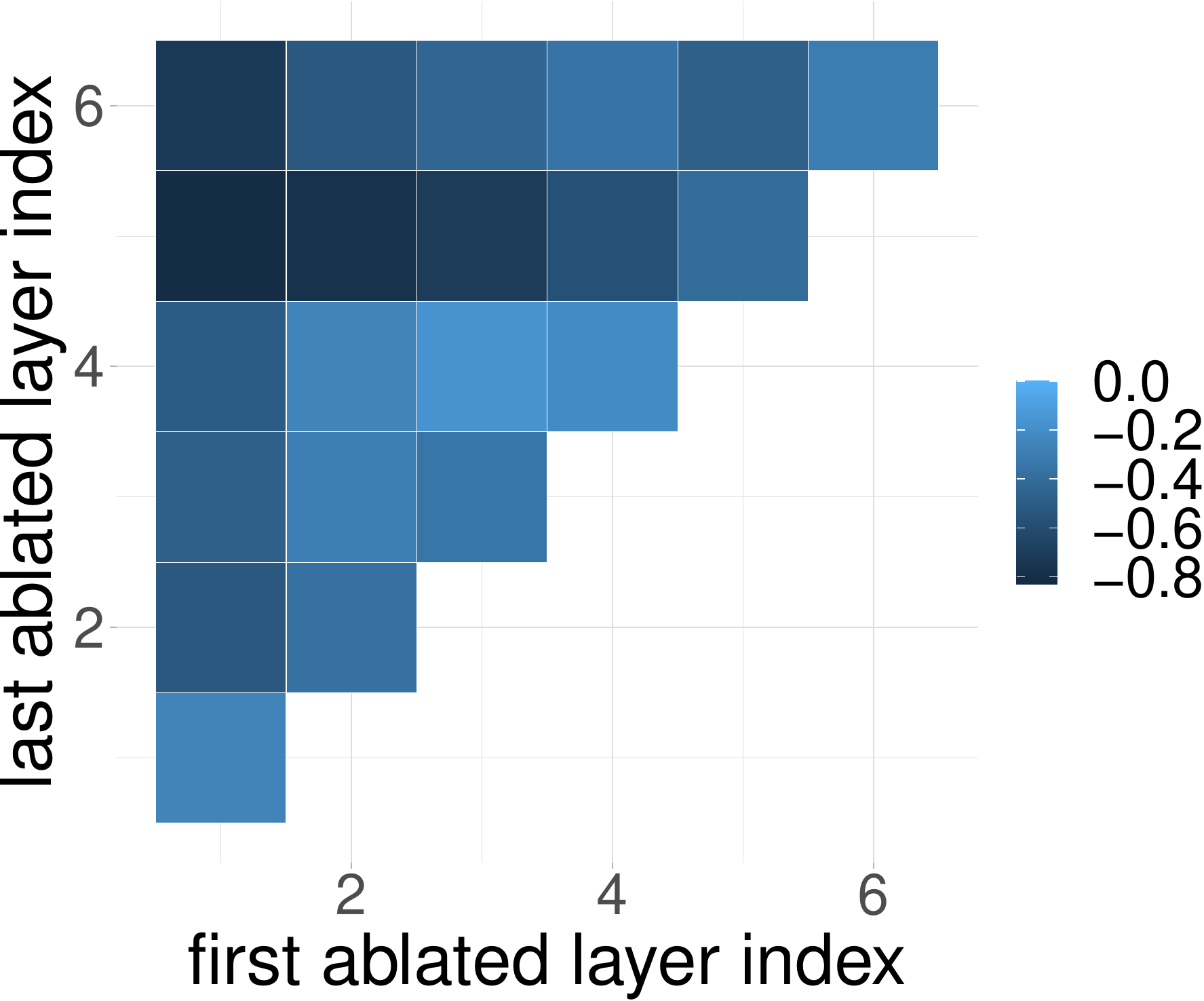}
        \caption[]%
        {{\small decoder zh2en}}    
        %\label{fig:mean and std of net44}
    \end{subfigure}
    \caption[Ablation study ]
    {Performance of \mymethod with ablated layers. The heat map shows the relative performance decrease as continuous layer spans are removed directly from the \mymethod model. The $x$ and $y$ axis indicate the index number of the first and the last layer removed. The layer adapters from the higher layer of the decoder side exert the most impact on the model when they are disabled.} 
    \label{fig:removeAdapter}
\end{figure}

We further examine these adapters' impact by re-evaluating the trained model with certain adapters from continuous layer spans removed, and we illustrate the BLEU score drop on Fig.~\ref{fig:removeAdapter}. 
%The drop is consistent with the trend of the L2-norm ratio:
We find that removing the adapters on the decoder side raises a greater performance decline, consistent with the trend of the L2-norm ratio.
Adapters of the \textit{en}$\rightarrow$\textit{zh} are more crucial to the multilingual models, which is in line with our experiments that adapters are more important for $\textit{en}\rightarrow\textit{any}$ directions~(Table~\ref{Tab:overall}). 
In addition, the decoder's upper layers of the \mymethod layer adapter have bigger impacts on the performance, consistent with \citet{houlsby2019parameter}'s findings on adapters for BERT model.
% In addition, intuitively, the more adapters are removed, the more the model's performance decreases.

\begin{figure}[]
\centering
\includegraphics[width=1\linewidth]{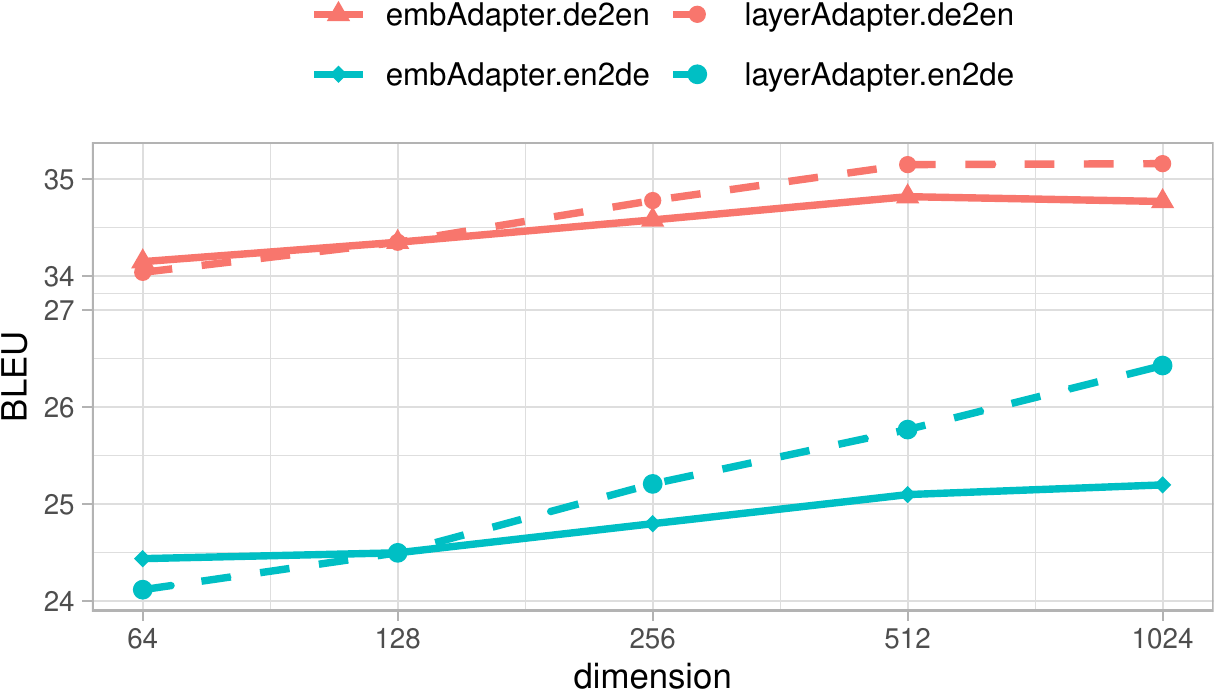}
\caption{BLEU score vs. \mymethod dimension on IWSLT en-de. 
The dashed line indicates the layer adapter, the solid line indicates the embedding adapter, and the color represents translation direction.
The BLEU score is more sensitive to the dimension of layer adapters.
Note part of the y-axis is truncated. 
}
\label{fig:bleu_dim_}
\end{figure}

\subsection{Parameter-Performance Trade-off}
The bottle-neck adapter design utilizes a small middle layer to control parameter efficiency~\cite{houlsby2019parameter}, while empirically, a larger layer dimension improves the performance via increased capacity. We explore the parameter-performance trade-off by varying the dimension of adapters and illustrate the BLEU score over different layer adapter and embedding adapter size in Fig.~\ref{fig:bleu_dim_} with different color, respectively.
Here we discuss the trade-off on \textit{en}$\leftrightarrow$\textit{de} directions of IWSLT. 
The plot shows that the dimension of the layer adapters has a significant impact to the performance: as the dimension doubles, the BLEU increases 0.58 and 0.28 by an average in \textit{en}$\rightarrow$\textit{de} and  \textit{de}$\rightarrow$\textit{en} directions respectively. In comparison, changing the embedding adapters' dimension impacts less on the final performance. When the dimension is large already, expanding the size hardly increases the BLEU score. 
Considering that the parameter amounts of \mymethod are small compared to the base model, and the final performance is sensitive to the layer adapter's dimension, we regard expanding dimension to be regarded as a simple and effective way to improve the performance of \mymethod.

% \begin{figure}[!h]
% \centering
% \includegraphics[width=1\linewidth]{figure/iwslt_bleu_dim-crop.pdf}
% \caption{BLEU score vs. dimension of layer adapter on IWSLT en-de. The dotted line indicates the results of bilingual baseline.\yaoming{alternative fig in Appendix}}
% \label{fig:bleu_layerAda}
% \end{figure}

% \begin{figure}[!h]
% \centering
% \includegraphics[width=1\linewidth]{figure/iwslt_bleu_emb_dim-crop.pdf}
% \caption{BLEU score vs. dimension of embedding adapter on IWSLT en-de. The dotted line indicates the results of bilingual baseline. }
% \label{fig:bleu_embAda}
% \end{figure}

\section{Conclusion}
% zym TODO refine

This work analyzes the performance degradation problem in multilingual NMT systems and decomposes into multilingual  embedding deficiency and multilingual interference effects. 
We then propose a novel framework to deal with degradation, named Counter-interference Adapter for Multilingual machine translation (\mymethod). \mymethod alleviates two issues above respectively by introducing two kinds of adapters. 

We validate the effectiveness of \mymethod on three multilingual translation datasets, where the results show that \mymethod improves the performance of the multilingual NMT model on various translation directions. 
The experiments also demonstrate that \mymethod variants outperform several strong baselines, approving our analysis and framework design. Furthermore, we investigate the behavior and utility of each component via empirical studies.

%\yaoming{summarize}
%Considering that the amount of parameters required by \mymethod is small compared to the base model, we 

% In this work, we analyze the performance degradation of multilingual NMT models from layer level. 
% We believe that the incapacity of multilingual models induces \textit{multilingual bias} to each sub-layer compared to bilingual baselines. To alleviate the issue, we then propose a new framework named \mymethod. The framework enhances the frozen multilingual base model on specific language pairs via two kinds of adapters: (i) layer adapters connected in parallel with base sub-layers, (ii) embedding adapters simulating the fine-tuned embeddings. 

% *(based on which we give practical suggestions about the training of the proposed framework.)*

\bibliography{anthology,acl2020}
\bibliographystyle{acl_natbib}

\clearpage

\section*{Appendix}

\subsection{Detailed Dataset Statistics}

We give the detailed statistics about the dataset used in Sec.~\ref{Sec:Exp}

\paragraph{IWSLT.} We almost follow the \citet{tan2018multilingual}'s script\footnote {\url{https://github.com/RayeRen/multilingual-kd-pytorch/blob/master/data/iwslt/raw/prepare-iwslt14.sh}} except that we removed their lowercase option. We collect training sets of 8 languages ranging from the year 2014 to 2016 and use the official valid/test set. We list the number of samples in training set in Table~\ref{iwslt_1}.

\begin{table}[h]
\centering
\resizebox{\columnwidth}{!}
{
\begin{tabular}{cccccccc}
\hline
ar   & de   & es   & fa  & he   & it   & nl   & pl   \\ \hline
140k & 160k & 169k & 89k & 144k & 167k & 153k & 128k \\ \hline
\end{tabular}
}
\caption{The number of training pairs of IWSLT}
\label{iwslt_1}
\end{table}

\paragraph{OPUS-100.} 
We collect data from \citet{zhang-etal-2020-improving}'s release\footnote{
\url{https://object.pouta.csc.fi/OPUS-100/v1.0/opus-100-corpus-v1.0.tar.gz}
}, and use its official valid/test set.
Among the 20 language pairs selected, 17 have 1 million training samples while three language pairs are of low resources, which are be(67k), nb(142k), and af(275k).

\paragraph{WMT.} 
We list the year of the training, valid and test set of each language in Table~\ref{Table:wmt_year}. Table~\ref{wmt_1} illustrate the number of samples in the training set.

\begin{table}[h]
\resizebox{\columnwidth}{!}
{
\begin{tabular}{c|ccccc}
\hline
         & es   & de   & ru   & et   & lv   \\ \hline
training & 2013 & 2016 & 2016 & 2018 & 2017 \\
valid    & 2012 & 2013 & 2019 & OPUS & OPUS \\
test     & 2013 & 2014 & 2020 & 2018 & 2017 \\ \hline
\end{tabular}
}
\caption{The year of training, valid test set of WMT datasets. The OPUS denotes we use the valid set from OPUS-100 dataset.}
\label{Table:wmt_year}
\end{table}

\begin{table}[h]
\centering
\resizebox{0.8\columnwidth}{!}
{
\begin{tabular}{cccccccc}
\hline
 es   & de   & ru   & et   & lv   \\ \hline  
15.18M & 4.56M & 2.59M & 2.18M & 4.51M \\ \hline
\end{tabular}
}
\caption{The number of training pairs of WMT}
\label{wmt_1}
\end{table}

\subsection{Detailed Experiment Results of Ablation Study and Model Variants}

To further study the efficacy of each component in \mymethod, we propose two variants as ablation study:

\noindent \textbf{\mymethod-layer}: This variant keeps only the layer adapter and removes all the embedding adapter.

\noindent \textbf{\mymethod-basic}: Besides removing all embedding adapters, this variant introduces only one layer adapter for each attention block. We design this variant to compare with Serial~\cite{bapna2019simple} under a similar amount of parameters to determine the effectiveness of our parallel connection further.

We present the detailed results on two \mymethod variants on Table~\ref{Tab:iwslt_main},~\ref{Tab:opus_ext} and \ref{Tab:wmt_ext} respectively. 
We give two major findings of ablation study: 
\begin{inparaenum}[\it a)]
\item  Compared with \mymethod, \mymethod-layer suffers from degradation in most language pairs (51 out of 66), especially in low-resource corpora(IWSLT). It further shows the effectiveness of the embedding adapter. \item With the same amount of parameters, \mymethod-basic surpass Serial in both overall performance and the number of improved pairs among all datasets. It shows that parallel is a more suitable adapter connection schema for multilingual machine translation. 
\end{inparaenum}

\subsection{The Effectiveness of Parallel Connection on Other Adapter Model}

As mentioned in Section~\ref{Sec:Exp}, we also substitute another adapter-based model with parallel connection methods to illustrate our proposed parallel connection is more suitable for multilingual machine translation. We re-implement \citet{philip2020language}'s work, where they proposed monolingual adapter which is specific to the source/target language other than the translation pair. We present the results of their serial connection and our parallel variants in Tab.~\ref{Tab:iwslt_main} as Mono-Serial and Mono-Parallel. 

We find parallel connection boosts the Mono adapter in 15 out of 16 language pairs in the IWSLT data set, the improvement is even more prominent in the \textit{en} $\rightarrow$ \textit{any} ones. The results further prove that our parallel layer adapter can provide improvement for all multilingual adapter models.

% \begin{table}[t]
% \centering
% \resizebox{1\columnwidth}{!}
% {
% \begin{tabular}{c|cccccc}
% \hline
%                               & \multicolumn{2}{c|}{en-IWSLT} & \multicolumn{2}{c|}{en-OPUS100} & \multicolumn{2}{c}{en-WMT}   \\ \hline
%                               & $\leftarrow$  & $\rightarrow$ & $\leftarrow$   & $\rightarrow$  & $\leftarrow$ & $\rightarrow$ \\ \hline
% Serial                         & 31.63         & 21.37         & 36.44          & 31.26          & 30.69        & 26.92         \\ \hline
% \mymethod-basic & 31.75         & 22.06         & 36.78          & 31.96          & 31.12        & 27.34         \\
% \mymethod-layer & 32.31         & 22.13         & 36.89          & 32.05          & 31.14        & 27.66         \\ \hline
% %WR\%-basic       & 87.5         & 100         & 25          & 75          & 60        & 80         \\
% %WR\%-layer       & 87.5         & 100         & 25          & 75          & 60        & 80         \\ \hline
% \end{tabular}
% }
% \caption{Overall Performance for two variants and of ablation study compared to Serial. The overall score is the arithmetic mean of the case-sensitive tokenized BLEU score of the test set of all languages. }\label{Tab:overall_variants}
% \end{table}

\begin{table*}[h]
\centering
\resizebox{2.1\columnwidth}{!}
{
\begin{tabular}{c|c|cccccccccccccccc}
\hline
             & params & \multicolumn{2}{c}{en-ar}      & \multicolumn{2}{c}{{ en-de}} & \multicolumn{2}{c}{{ en-es}} & \multicolumn{2}{c}{en-fa}      & \multicolumn{2}{c}{en-he}      & \multicolumn{2}{c}{en-it}      & \multicolumn{2}{c}{en-nl}      & \multicolumn{2}{c}{en-pl}      \\ \hline
             &          & $\leftarrow$ & $\rightarrow$ & $\leftarrow$                      & $\rightarrow$                      & $\leftarrow$                     & $\rightarrow$                    & $\leftarrow$ & $\rightarrow$ & $\leftarrow$ & $\rightarrow$ & $\leftarrow$ & $\rightarrow$ & $\leftarrow$ & $\rightarrow$ & $\leftarrow$ & $\rightarrow$ \\ \hline
Bilingual    & 20M$\times N$   & 31.35        & 13.82         & 34.65        & 25.81         & 40.62        & 35.41         & 22.76        & 11.90         & 36.78        & 25.16         & 34.13        & 30.72         & 36.65        & 30.18         & 22.44        & 14.51         \\ \hline
Multilingual & 20M        & 30.01        & 12.09         & 33.54        & 22.40         & 39.56        & 32.18         & 22.01        & 11.15         & 34.54        & 20.35         & 33.98        & 27.88         & 35.72        & 27.36         & 23.15        & 12.07         \\
KD                                & 0        & 30.03        & 12.31         & 32.03        & 22.55         & 38.06        & 30.78         & 22.12        & 10.51         & 33.32        & 20.69         & 32.93        & 25.65         & 35.84        & 27.15         & 23.45        & 12.80          \\
Serial       & 265k     & 30.05        & 12.52         & 33.65        & 23.48         & 39.57        & 33.02         & 22.21        & 11.90         & 34.59        & 20.37         & 33.54        & 28.80         & 35.86        & 28.19         & 23.57        & 12.66         \\ \hline
\mymethod-basic            & 20M $+$ 264k$\times N$     & 30.11        & 12.48         & 33.46        & 24.12         & 39.74        & 33.73         & 22.22        & 12.48         & 34.82        & 22.28         & 34.14        & 29.32         & 36.01        & 28.90          & 23.50         & 13.22         \\
\mymethod-layer   & 20M $+$ 528k$\times N$     & 30.51          & 12.99          & 34.29          & 24.02         & \textbf{40.48} & 33.49          & 22.63          & 12.59          & \textbf{35.20} & 22.55         & \textbf{34.73} & 29.34          & 36.55          & 28.80           & \textbf{24.13} & 13.24          \\
\mymethod         &20M $+$ 660k $\times N$      & \textbf{30.74} & \textbf{13.31} & \textbf{34.35} & \textbf{24.5} & 40.13          & \textbf{34.16} & \textbf{23.64} & \textbf{12.73} & 35.12         & \textbf{22.7} & 34.50           & \textbf{29.74} & \textbf{36.67} & \textbf{29.33} & 23.96          & \textbf{13.41} \\
\hline \hline 
Mono-Serial   & 20M $+$ 132k$\times L$                       & 30.25                           & 11.14                           & 33.97                           & 22.66                          & 39.27                           & 32.26                           & 22.46                           & 10.79                           & 33.86                           & 20.76                          & 33.53                           & 28.11                           & 36.45                           & 27.46 & 23.04 & 12.14  \\
Mono-Parallel & 20M $+$ 132k$\times L$                      & 30.69                           & 12.40                           & 33.93                           & 23.42                          & 39.83                           & 33.15                           & 22.74                           & 11.56                           & 34.35                           & 20.82                          & 34.17                           & 28.84                           & 36.50                           & 28.30 & 23.68 & 12.61  \\ \hline

\end{tabular}
}
\caption{Full results of Case-sensitive tokenized BLEU on the IWSLT dataset. params(M) represents the total number of parameters required for each model in million, where $N$ is the number of language pairs, and $L$ is the total number of languages.
}
\label{Tab:iwslt_main}
\end{table*}

\begin{table*}[h]
\centering
\scriptsize
\resizebox{2\columnwidth}{!}
{
\begin{tabular}{c|cccccccccccccc}
\hline
           & \multicolumn{2}{c|}{$\Delta$params} & \multicolumn{2}{c}{en-ar}    & \multicolumn{2}{c}{en-fa}    & \multicolumn{2}{c}{en-de}    & \multicolumn{2}{c}{en-nl}    & \multicolumn{2}{c}{en-af}    & \multicolumn{2}{c}{en-da}    \\
           & \multicolumn{2}{c|}{}               & $\leftarrow$ & $\rightarrow$ & $\leftarrow$ & $\rightarrow$ & $\leftarrow$ & $\rightarrow$ & $\leftarrow$ & $\rightarrow$ & $\leftarrow$ & $\rightarrow$ & $\leftarrow$ & $\rightarrow$ \\ \hline
\mymethod-basic & \multicolumn{2}{c|}{12.6M}          & 40.94        & 25.21         & 25.58        & 10.94         & 35.79        & 32.09         & 34.13        & 30.38         & 56.45        & 52.55         & 39.12        & 38.13         \\
\mymethod-layer & \multicolumn{2}{c|}{25.2M}          & 41.11        & 25.13         & 25.83        & 10.96         & 35.92        & 32.27         & 34.23        & 30.33         & 56.18        & 52.46         & 39.19        & 38.21         \\ \hline
           & \multicolumn{2}{c}{en-no}           & \multicolumn{2}{c}{en-sv}    & \multicolumn{2}{c}{en-nb}    & \multicolumn{2}{c}{en-es}    & \multicolumn{2}{c}{en-pt}    & \multicolumn{2}{c}{en-fr}    & \multicolumn{2}{c}{en-it}    \\
           & $\leftarrow$     & $\rightarrow$    & $\leftarrow$ & $\rightarrow$ & $\leftarrow$ & $\rightarrow$ & $\leftarrow$ & $\rightarrow$ & $\leftarrow$ & $\rightarrow$ & $\leftarrow$ & $\rightarrow$ & $\leftarrow$ & $\rightarrow$ \\ \hline
\mymethod-basic & 28.63            & 31.03            & 33.85        & 34.59         & 45.61        & 39.88         & 41.86        & 39.65         & 39.36        & 34.96         & 36.35        & 36.28         & 37.95        & 33.41         \\
\mymethod-layer & 28.58            & 31.08            & 33.98        & 34.82         & 45.89        & 40.11         & 41.8         & 39.72         & 39.63        & 34.75         & 36.37        & 36.72         & 37.81        & 33.11         \\ \hline
           & \multicolumn{2}{c}{en-pl}           & \multicolumn{2}{c}{en-cs}    & \multicolumn{2}{c}{en-sk}    & \multicolumn{2}{c}{en-ru}    & \multicolumn{2}{c}{en-uk}    & \multicolumn{2}{c}{en-be}    & \multicolumn{2}{c}{en-zh}    \\
           & $\leftarrow$     & $\rightarrow$    & $\leftarrow$ & $\rightarrow$ & $\leftarrow$ & $\rightarrow$ & $\leftarrow$ & $\rightarrow$ & $\leftarrow$ & $\rightarrow$ & $\leftarrow$ & $\rightarrow$ & $\leftarrow$ & $\rightarrow$ \\ \hline
\mymethod-basic & 28.79            & 23.84            & 36.39        & 29.05         & 41.18        & 34.08         & 35.96        & 31.82         & 29.88        & 19.5          & 25.58        & 20.19         & 42.29        & 41.74         \\
\mymethod-layer & 28.46            & 24.44            & 37.06        & 29.22         & 41.14        & 33.92         & 36.21        & 31.77         & 29.86        & 19.41         & 25.94        & 20.85         & 42.51        & 41.62         \\ \hline
\end{tabular}
}
\caption{Case-sensitive tokenized BLEU score on the OPUS-100 dataset of two model variants.}
\label{Tab:opus_ext}
\end{table*}

\begin{table*}[!h]
\centering
\scriptsize
\resizebox{1.8\columnwidth}{!}
{
\begin{tabular}{c|c|cccccccccc}
\hline
             & $\Delta$params & \multicolumn{2}{c}{en-es}    & \multicolumn{2}{c}{en-de}    & \multicolumn{2}{c}{en-et}    & \multicolumn{2}{c}{en-ru}    & \multicolumn{2}{c}{en-lv}    \\
             &                & $\leftarrow$ & $\rightarrow$ & $\leftarrow$ & $\rightarrow$ & $\leftarrow$ & $\rightarrow$ & $\leftarrow$ & $\rightarrow$ & $\leftarrow$ & $\rightarrow$ \\ \hline
\mymethod-basic   & 12.6M          & 35.79 & 34.25          & 33.62          & 27.36          & 28.11          & 20.68          & 36.62         & 35.96          & 21.45          & 18.45          \\
\mymethod-layer   & 25.2M          & 35.70           & 34.40           & 33.65 & 27.85 & 28.21          & 20.91          & 36.55         & 36.22 & 21.58          & 18.91          \\ \hline
\end{tabular}
}
\caption{
Case-sensitive tokenized BLEU score on the WMT dataset of two model variants.
}
\label{Tab:wmt_ext}
\end{table*}

\subsection{Case Study}

We also conduct qualitative analysis by case study. We invite two German speakers to compare the translation of a news report from the WMT test set. The contestants are generated by \mymethod, Serial and vanilla multilingual. 
We find German translation of \mymethod are more favored by human annotators for the following reasons: 
\begin{inparaenum}[\it a)]
\item  The sentence tense and clause pattern fit the original sentence; 
\item \mymethod tend to use set phrases instead of simple expressions; 
\item \mymethod better captures the relationship between modifiers and the subjects.
\end{inparaenum} 

We list a sample translation in Tab~\ref{Tab:case}.

\begin{table}[]
\scriptsize
\begin{tabular}{c|p{6cm}}
\hline
Source & The Kluser lights protect cyclists, as well as those travelling by bus and the residents of Bergle. \\ \hline
Multilingual &   Die Kluser-Lichter schützen Radfahrer, Busfahrer und Bergleiter.                      \\ \hline
Serial &   Die Lichter von Kluser schützen Radfahrer, aber auch Busreisende und die Bewohner von Bergle.                      \\ \hline
\mymethod    &   Die Kluser-Leuchten schützen Radfahrer, Busfahrer und Einwohner von Bergle.  
\\ \hline
Human  & Die Kluser-Ampel sichere sowohl Radfahrer als auch Busfahrgäste und die Bergle-Bewohner.            \\ \hline
\end{tabular}
\caption{Sampled translation of \mymethod and baselines. We also list human reference here for comparison. The sample shows that the \mymethod's translation is more accurate and smooth. }
\label{Tab:case}
\end{table}

% \begin{table*}[h]
% \centering
% \scriptsize
% \caption{Case-sensitive tokenized BLEU score on the WMT dataset of two model variants. $\Delta$params represents the additional parameters required for each new language pair.}
% \resizebox{2\columnwidth}{!}
% {
% \begin{tabular}{c|c|cccccccccc}
% \hline
%           & $\Delta$params & \multicolumn{2}{c}{en-es}    & \multicolumn{2}{c}{en-de}    & \multicolumn{2}{c}{en-et}    & \multicolumn{2}{c}{en-ru}    & \multicolumn{2}{c}{en-lv}    \\
%           &                & $\leftarrow$ & $\rightarrow$ & $\leftarrow$ & $\rightarrow$ & $\leftarrow$ & $\rightarrow$ & $\leftarrow$ & $\rightarrow$ & $\leftarrow$ & $\rightarrow$ \\ \hline
% \mymethod-basic & 12.6M          & 35.79        & 34.25         & 33.62        & 27.36         & 28.11        & 20.68         & 36.62        & 35.96         & 21.45        & 18.45         \\
% \mymethod-layer      & 25.2M          & 35.7         & 34.4          & 33.65        & 27.85         & 28.21        & 20.91         & 36.55        & 36.22         & 21.58        & 18.91         \\ \hline
% \end{tabular}
% }
% \end{table*}

\clearpage
\end{CJK}

\end{document}